%% file: main.tex
\documentclass{article} 

\usepackage{microtype}
\usepackage{graphicx}
\usepackage{subfigure}
\usepackage{booktabs} 

\usepackage{hyperref}

\usepackage[preprint]{icml2024}

\input{math_commands.tex}

\usepackage{tablefootnote}
\usepackage{hyperref}
\usepackage{url}

\usepackage[utf8]{inputenc} 
\usepackage[T1]{fontenc}    
\usepackage{graphicx}
\usepackage{booktabs}       
\usepackage{amsfonts}       
\usepackage{nicefrac}       
\usepackage{microtype}      
\usepackage{xcolor}         
\usepackage{multirow}
\usepackage[noend]{algpseudocode}
\usepackage{amsbsy}
\usepackage{amssymb}
\usepackage{amsmath}

\newcommand{\electricity}{\texttt{electricity}}
\newcommand{\traffic}{\texttt{traffic}}
\newcommand{\ettm}{\texttt{ETTm2}}
\newcommand{\weather}{\texttt{weather}}

\newcommand{\method}{\texttt{tsGT}}

\begin{document}

\twocolumn[
\icmltitle{\method{}: Stochastic Time Series Modeling With Transformer}

\icmlsetsymbol{equal}{*}

\begin{icmlauthorlist}
\icmlauthor{Łukasz Kuciński}{ideas,impan,mimuw}
\icmlauthor{Witold Drzewakowski}{ideas,mimuw}
\icmlauthor{Mateusz Olko}{ideas,uw}
\icmlauthor{Piotr Kozakowski}{uw}
\icmlauthor{Łukasz Maziarka}{uj}
\icmlauthor{Marta Emilia Nowakowska}{mimuw}
\icmlauthor{Łukasz Kaiser}{openai}
\icmlauthor{Piotr Miłoś}{ideas,impan,mimuw,deepsense}
\end{icmlauthorlist}

\icmlaffiliation{ideas}{IDEAS NCBR}
\icmlaffiliation{impan}{Institute of Mathematics, Polish Academy of Sciences}
\icmlaffiliation{uj}{Jagiellonian University, Faculty of Mathematics and Computer Science}
\icmlaffiliation{uw}{University of Warsaw, Doctoral School of Exact and Natural Science}
\icmlaffiliation{mimuw}{University of Warsaw, Faculty of Mathematics, Informatics and Mechanics}
\icmlaffiliation{openai}{OpenAI}
\icmlaffiliation{deepsense}{deepsense.ai}

\icmlcorrespondingauthor{Łukasz Kuciński}{lukasz.kucinski@ideas-ncbr.pl}

\icmlkeywords{Machine Learning, ICML}

\vskip 0.3in
]

\printAffiliationsAndNotice{} 

\begin{abstract}

Time series methods are of fundamental importance in virtually any field of science that deals with temporally structured data. Recently, there has been a surge of deterministic transformer models with time series-specific architectural biases. In this paper, we go in a different direction by introducing \method{}, a stochastic time series model built on a general-purpose transformer architecture. We focus on using a well-known and theoretically justified rolling window backtesting and evaluation protocol. We show that \method{} outperforms the state-of-the-art models on MAD and RMSE, and surpasses its stochastic peers on QL and CRPS, on four commonly used datasets. We complement these results with a detailed analysis of \method{}'s ability to model the data distribution and predict marginal quantile values.

\end{abstract}

\section{Introduction}

Researchers and practitioners use data in the form of time series to predict and analyze future events, estimate relationships, and make informed decisions across a diverse range of scientific fields. With applications spanning medicine, finance, economics, meteorology, astronomy, transportation, and manufacturing, the significance of time series methods cannot be overstated.  This widespread importance is further underscored by the Nobel Memorial Prize in Economic Sciences awarded to breakthrough discoveries in the field \citep{nobelprize2003}.

The generative mechanism behind time series data is often modeled as an autoregressive stochastic process \citep{hamilton2020time}. These models encompass a broad range of approaches, such as classical econometrics techniques like ARMA \citep{whittle1951hypothesis} and ARCH models \citep{engle1982autoregressive}, or recurrent neural networks \citep{hochreiter1997long} and  transformers \citep{vaswani2017attention} in deep learning. In particular, the last class of models emerged as a powerful architecture that fuels many of the recent advances in machine learning, including text or code generation  \citep{brown2020language,chen2021evaluating}, solving mathematical problems \citep{lewkowycz2022solving,polu2022formal,mikula2023magnushammer}, planning \citep{czechowski2021subgoal, zawalski2023fast}, hyperparameter tuning \citep{kozakowski2020forecasting, chen2022towards}, EEG analysis \citep{kostas2021bendr}.

Transformers have also made their way to the time series domain \citep{li2019enhancing, woo2022etsformer,zhou2022fedformer,wu2021autoformer,zhou2021informer,wen2023transformers}. Interestingly, contemporary models of this type are deterministic and are evaluated using pointwise metrics, such as mean absolute deviation (MAD) or root mean square error (RMSE), within a single time window. Such models have many benefits, including simplicity, ease of interpretation, and computation efficiency. Having said that, time series are inherently stochastic and we believe they should be modeled as such. In particular, the models should be useful in applications such as scenario analysis, uncertainty assessment, or accurate data description. In this paper, we aim to make progress in this direction, emphasizing general-purpose architectures and well-established time series evaluation protocols.

We present \method{}, a generative decoder-only transformer stochastic time series model. A key feature of \method{} is its stochasticity, which we obtain effortlessly due to the transformer distributional output. Additionally, the model is built upon a general-purpose architecture, without featuring any domain-specific inductive biases typically required in neural network design, input processing, or loss function formulation. This reduces the costs of training due to decreased number of dataset specific hyperparameters and facilitates effective scaling with data, computation, and model size \citep{sutton2019bitter, kaplan2020scaling, hoffmann2022training}. This approach has led to a considerable increase in performance on MAD and RMSE, with \method{} outperforming contemporary methods. As a byproduct, we show that \method{} is not permutationally invariant, addressing a prominent critique of contemporary transformer models \cite{Zeng2022AreTE}.

Similarly, \method{} surpasses stochastic peers on quantile metrics, the quantile loss (QL), and continuous ranked probability score (CRPS) on all datasets. This result is further reinforced by a detailed analysis of \method{} ability to model data distribution, employing a quantile backtesting procedure as outlined by \citep{mcneil2015quantitative}. Importantly, we focus on the model's predictive capabilities, employing a rolling window analysis, which is a well-known time-series procedure that is more robust to outliers and allows us to assess the methods’ stability over time. \looseness-1

Our contributions are as follows:
\begin{itemize} 
    \item We propose \method{}, a general-purpose transformer stochastic time series model. Additionally, we focus on using a rolling window evaluation protocol and backtesting the predictive performance.

    \item We show that in terms of MAD and RMSE, \method{} outperforms the contemporary models on all considered datasets. 
    
    \item We rigorously assess \method{}'s predictive performance. In particular, we show that \method{} outperforms stochastic baselines on QL and CRPS, analyze \method{}'s ability to model data distribution, and demonstrate that it is not permutationally invariant.
\end{itemize}

\section{Related work}\label{sec:related_work}

The classical treatment of time series modeling can be found in \citep{hamilton2020time,brockwell2009time,HyndmanAthanasopoulos2021}. Historically, much of the progress in the field has been driven by financial and risk management applications \citep{nobelprize2003,zivot2006modeling, tsay2010analysis,mcneil2015quantitative}. More recently, deep learning methods \citep{goodfellow2016deep}, such as RNNs and Transformers have made their way into the field \citep{salinas2020deepar, li2019enhancing}. Transformers are strong sequence-to-sequence architectures, and their large-scale variants are known as large language models (LLMs). There is an overwhelming abundance of LLMs, including GPT variants  \citep{radford2019language, brown2020language, OpenAI2023GPT4TR}, PaLM \citep{chowdhery2022palm}, LLaMA \citep{Touvron2023LLaMAOA}, or Gemini \citep{team2023gemini}. See \citep{liang2022holistic, zhao2023survey, yang2023harnessing} for recent surveys on the topic. Adaptation of LLMs to time series tasks has been explored in \citep{gruver2023large, wu2023bloomberggpt}. We employ architectural choices found to be useful in LLMs, including rotary embeddings \citep{su2021roformer} and GELU activation function \citep{hendrycks2016gaussian}. 

Following the popularity of transformers, there has been a surge of transformer-inspired time series models; see \citep{wen2023transformers} for a recent survey. Recent deterministic models include ETSformer \citep{woo2022etsformer}, FEDformer \citep{zhou2022fedformer},  Autoformer \citep{wu2021autoformer}, Informer \citep{zhou2021informer}, PatchTST \citep{nie2022time}, Crossformer \citep{zhang2023crossformer}. Some of these models were criticized by \cite{Zeng2022AreTE}, as they were shown to be permutationally invariant in some evaluation settings and to fall short of a simple MLP-based model (DLinear). Stochastic transformer time series models include ConvTrans \citep{li2019enhancing},  Temporal Fusion Transformer \citep{lim2021temporal}, IQN-Transformer \citep{gouttes2021probabilistic},  FQFormer \citep{jawed2022fqformer}. Other all-MLP models for time series include N-BEATS \citep{oreshkin2020nbeats}, N-HiTS \citep{challu2022nhits} and TSMixer \citep{chen2023tsmixer}.

There are multiple ways of handling real-valued data in the literature. Typically, the raw time series values are normalized before further processing, e.g., by scaling to a certain interval \citep{li2019enhancing} or centered and standardized \citep{nie2022time}. After that, the data is tokenized, via bucketing \citep{chen2021decision, janner2021offline, chen2022towards, reed2022generalist} or $k$-means \citep{shafiullah2022behavior}, or directly embedded as a multivariate vector e.g., by introducing covariates \citep{li2019enhancing}, modeling multivariate time series \citep{zhou2021informer, wu2021autoformer, zhou2022fedformer, woo2022etsformer}, or patching  \citep{nie2022time, zhou2023power}. \method{} handles preprocessing of real-valued data in three steps: (a) the values are normalized to $[0,1]$ interval, similar to \citep{li2019enhancing}, (b) the values are quantized into a fixed precision, and (c) the values are split into digits. The last step is similar to \citep{nogueira2021investigating, chowdhery2022palm}, where tokenization of string representation of numbers into digits was used to improve parsing simple text-based math problems. Steps (b) and (c) can also be viewed as a form of vector quantization \citep{gray1984vector, lendasse2005vector, oord2017neural}.

The contemporary time series methods mentioned above are trained and evaluated on a fixed time window. In this paper, we perform a rolling-window analysis, i.e., we train and evaluate models on multiple overlapping windows \citep{mcneil2000estimation, gonzalez2004forecasting, mcneil2015quantitative}. This allows assessing the methods' stability over time and their predictive performance via backtesting procedure \citep{mcneil2015quantitative}. Finally, we leave the efficient transformer implementation out of the paper's scope, since it is a challenge studied intensively across the deep learning field, see e.g., \citep{tay2022efficient}. This allows us to focus on modeling complex stochastic properties of continuous data using a plug-and-play design, which facilitates the integration of efficiency-related innovation in the field.

\section{Time series model: \method{}}\label{sec:method}

\paragraph{Time series notation}

We consider a time series which ends at time $t$ and has a length $T$, $\mathbb X_{t-T+1:t}=(X_{t-T+1}, \ldots, X_{t})$. Our goal is to predict the $H$ timesteps following $t$, i.e., $\mathbb X_{t+1:t+H}=(X_{t+1}, \ldots, X_{t+H})$. The raw data $\mathbb{X}_{t-T+1:t}$ forms an input to the model, and no additional information is provided (such as series id, timestamp, or other exogenous factors).

\paragraph{Transformer backbone}

We use a decoder-only transformer~\citep{vaswani2017attention}, with pre-normalization~\citep{radford2019language}, rotary embeddings~\citep{su2021roformer}, and GELU activation function~\citep{hendrycks2016gaussian}. These choices are inspired by recent successful architectures like GPT-3 \citep{brown2020language} or LLaMA \citep{Touvron2023LLaMAOA}. For architectural details, see Appendix \ref{app:architecture}-\ref{app:hyperparameters}. Notice that we do not use time series-specific architectural modifications, such as 1D convolutions \citep{li2019enhancing}, frequency domain  \citep{wu2021autoformer, zhou2022fedformer}, seasonal-trend decomposition \citep{woo2022etsformer}, or patching \citep{nie2022time}.

\paragraph{Input and output discretization}

Before the raw time series data are passed to the transformer backbone, they are tokenized. 
Tokenization is performed in three stages: (a) normalization of data, (b) quantization into fixed precision, and (c) discretization into digits.

During the normalization stage, the values are squeezed to interval $[0, 1]$, a procedure also found in \citep{salinas2020deepar,janner2021offline,chen2022towards,reed2022generalist}, see Appendix \ref{app:architecture}.

Next, each value $a\in[0,1]$ is tokenized to digits in a fixed precision $p\in \mathbb{N}_+$ in base $B>1$, via a function $\tau(a) = (\tau_1(a), \ldots, \tau_p(a))$, defined through an equation $\lfloor a\cdot B^p\rfloor =\sum_{k=1}^{p} \tau_k(a) \cdot B^{p-k}$. For example, with $p=3$ and $B=10$, a number $a=0.123$ is represented as three tokens \texttt{1}, \texttt{2}, \texttt{3}.

With a slight abuse of notation, we denote by $\tau(\mathbb{X})$ a series of tokens resulting from breaking a normalized time series $\mathbb{X}$ into digits (with a length of $\tau(\mathbb{X})$ being $p$ times larger than the length of $\mathbb{X}$). For each $k$, a model $g_\theta$ outputs a conditional distribution $g_\theta(\cdot|\tau_{1}, \ldots, \tau_{k})$ over the next $(k+1)$-th token (digit). If we denote $\tau'=(\tau_1',\ldots,\tau_{pT+pH}')=\tau(\mathbb{X}_{t-T+1:t+H})$, the conditional distribution of the predicted horizon is given by
\begin{equation*}
    \begin{split}
        g_\theta(
        \tau_{pT+1}', \ldots, \tau_{pT+pH}'
        |\tau_1',\ldots, \tau_{pT}') \\
        = \prod_{k=pt+1}^{pT+pH} g_\theta(\tau_k'|\tau_1',\ldots, \tau_{k-1}').
    \end{split}
\end{equation*}
\looseness=-1
The distribution of each real-valued number returned by the model is induced by the joint probability of $p$ consecutive digits and the mapping inversing the tokenization procedure. This joint probablity is recovered via a chain rule as a product of non-parametric conditional distributions returned by the model. This procedure gives the model probabilistic flexibility, allowing a range of complex and multi-modal distributions.

\paragraph{Loss function}

We use a standard transformer training objective of predicting the next token with a weighted cross-entropy loss:
\begin{equation}\label{eq:loss}
    \mathcal L(\theta) = \mathbb E_{
        \substack{\mathbb{X}\sim Data \\
        \tau' = \tau(\mathbb{X})}
    }
    \left[\sum_{k=1}^{pT}\beta^{(k-1)\% p} \log g_\theta(\tau_{k}'|\tau_1',\ldots, \tau_{k-1}')\right].
\end{equation}
Here $\beta\in(0, 1]$ is a hyperparameter discriminating between the significance of digits, and $k\%p$ denotes the remainder of $t$ divided by $p$.

\paragraph{Simulation}

For clarity of exposition, we omit discretization details in what follows, letting the reader infer specific details. The predictions are generated in an autoregressive manner using a standard Monte Carlo method. Assuming that for a time series of length $T$ ending at time $t$, $\mathbb{X}_{t-T+1:t}$, we have already generated $h$ steps of predictions, we generate the next value, $X_{t+h+1}$, as the sample from the model evaluated on $\mathbb{X}_{t-T+1:t+h}$. Consequently, we append $X_{t+h+1}$ to the sequence, resulting in $\mathbb{X}_{t-T+1:t+h+1}$, and the procedure is repeated. In this paper, we make $I=1024$ simulations of $H$ steps following $t$, for every considered $t$. 

\section{Experimental setup}\label{sec:experiments}

Below, we present the setup used for our experiments, postponing experimental results to the next section. In Section \ref{sec:main_results}, our main result shows the \method{} outperforms the baselines on all considered datasets and metrics. 

\paragraph{Transformer architecture}

\looseness=-1 
We set the precision $p=3$ and base $B=10$. During training, the raw time series input window is $T=256$, which translates to $pT=768$ tokens fed to the transformer. During inference, we set the prediction horizon $H=24$, and the raw time series input window is equal to $T=256-H=232$. In all our experiments, the transformer decoder comprises $6$ layers and $4$ self-attention heads, with an embedding dimension of $256$. Our models have $\approx 3.2$M trainable weights. For training, we use a batch size $16$, Adam optimizer, and scheduling the learning rate (multifactor approach with a constant of 0.03, linear warmup of 1000 steps, and square root decay). Additionally, we apply a weight decay with the rate of $10^{-5}$ and do not use early stopping. See Appendix \ref{app:hyperparameters} for more details.

\paragraph{Baseline models}

We include as baselines several contemporary transformer and non-transformer methods. Models from the former category include FEDformer \citep{zhou2022fedformer}, Informer \citep{zhou2021informer}, Autoformer \citep{wu2021autoformer}, ETSformer \citep{woo2022etsformer}, IQN-Transformer \citep{gouttes2021probabilistic}, PatchTST \citep{nie2022time}, and a family of stochastic models inspired by \cite{li2019enhancing}. This family includes variants with different parametric marginal distributions: Gaussian, Laplace, and t-student. Non-transformer models include DLinear \citep{Zeng2022AreTE} and TSMixer \citep{chen2023tsmixer}, representing linear and MLP-based approaches, respectively. Among the baselines, IQN-Transformer, Gaussian, Laplace, and t-student are stochastic, and other ones are deterministic. For the deterministic models, we use hyperparameters proposed by their open-source implementations. For the details on the hyperparameters used, see Appendix \ref{app:hyperparameters}. The models' approximate parameter count is as follows: FEDformer $18$M, Informer $12$M, Autoformer $11$M, and ETSformer $6$M. For more details on baselines, see Appendix \ref{app:stochastic_baselines}-\ref{app:baselines}, and Appendix \ref{app:infra} for computational infrastructure used.

\paragraph{Datasets}

To assess the performance of our method against the baseline models we use four real-life datasets, which are commonly used in the field: \electricity{}, \traffic{}, \ettm{}, and \weather{}, see \citet{wu2021autoformer}. The \electricity{} and \traffic{} datasets are rich in periodic patterns. \ettm{} introduces additional complexity, as it contains negative values, outliers, and displays unpredictable behavior. Finally, \weather{} dataset contains time series with varying scales, frequent outliers, and features a considerable number of zero values in certain series.

Every dataset comprises $S$ time series and each data point is time-stamped. The value $S$ is dataset-dependent, e.g., for \electricity{} $S=321$ and for \traffic{} $S=862$. The training window spans $365 \cdot 24=8760$ timesteps, immediately followed by the evaluation horizon of length $H=24$. See Appendix \ref{app:data} for further details.

\begin{table*}[t]
    \centering
    \caption{Performance of models in terms of error metrics MAD and RMSE (lower values are better). Each entry is computed using IQM and accompanied by a $90\%$ bootstrap confidence interval (in brackets)  computed over $W=100$ training windows and all time series.} 
    \label{tab:deterministic_preformance}
    \begin{tabular}{lllllll}
        \toprule
        metric & model & \electricity{} & \traffic{} & \ettm{} & \weather{} \\
        \midrule
        MAD & t-student$^{*}$  & 0.126\tiny{(0.125, 0.126)} & 0.273\tiny{(0.272, 0.274)} & 0.187\tiny{(0.172, 0.204)} & 0.125\tiny{(0.116, 0.134)}\\
         & Laplace$^{*}$ & 0.127\tiny{(0.126, 0.127)} & 0.272\tiny{(0.272, 0.273)} & 0.186\tiny{(0.171, 0.204)} & 0.125\tiny{(0.116, 0.134)}\\
         & Gaussian$^{*}$ & 0.125\tiny{(0.124, 0.125)} & 0.275\tiny{(0.274, 0.276)} & 0.182\tiny{(0.168, 0.199)} & 0.135\tiny{(0.126, 0.145)}\\
         & IQN$^{*}$ & 0.124\tiny{(0.123, 0.124)} & 0.349\tiny{(0.348, 0.351)} & 0.183\tiny{(0.169, 0.200)} & 0.129\tiny{(0.121, 0.137)}\\
         & Informer & 0.235\ \tiny{(0.233, 0.236)} & 0.195\ \tiny{(0.193, 0.196)} & 0.195\ \tiny{(0.194, 0.197)} & 0.224\ \tiny{(0.223, 0.226)}\\
         & ETSformer & 0.201\ \tiny{(0.200, 0.203)} & 0.192\ \tiny{(0.191, 0.193)} & 0.193\ \tiny{(0.192, 0.195)} & 0.206\ \tiny{(0.204, 0.207)}\\
         & FEDformer & 0.179\ \tiny{(0.178, 0.180)} & 0.162\ \tiny{(0.161, 0.164)} & 0.156\ \tiny{(0.154, 0.157)} & 0.180\ \tiny{(0.178, 0.181)}\\
         & Autoformer & 0.169\ \tiny{(0.167, 0.170)} & 0.161\ \tiny{(0.159, 0.162)} & 0.152\ \tiny{(0.151, 0.154)} & 0.173\ \tiny{(0.172, 0.174)}\\
         & DLinear & 0.065\ \tiny{(0.064, 0.065)} & 0.181\ \tiny{(0.180, 0.182)} & \textbf{0.094\ \tiny{(0.089, 0.101)}} & 0.111\ \tiny{(0.105, 0.118)}\\
         & PatchTST & 0.062\ \tiny{(0.062, 0.062)} & 0.152\ \tiny{(0.152, 0.153)} & \textbf{0.094\ \tiny{(0.089, 0.101)}} & 0.079\ \tiny{(0.073, 0.085)}\\
         & TSMixer & 0.063\ \tiny{(0.062, 0.063)} & 0.153\ \tiny{(0.152, 0.154)} & \textbf{0.090\ \tiny{(0.084, 0.096)}} & 0.087\ \tiny{(0.081, 0.092)}\\
         & \method{}$^{*}$ & \textbf{0.057\ \tiny{(0.057, 0.058)}} & \textbf{0.114\ \tiny{(0.114, 0.115)}} & \textbf{0.092\ \tiny{(0.086, 0.098)}} & \textbf{0.056\ \tiny{(0.052, 0.061)}}\\
        \midrule
         RMSE & t-student$^{*}$ & 0.164\tiny{(0.163, 0.165)} & 0.398\tiny{(0.397, 0.400)} & 0.215\tiny{(0.198, 0.235)} & 0.143\tiny{(0.133, 0.154)}\\
         & Laplace$^{*}$ & 0.165\tiny{(0.165, 0.166)} & 0.399\tiny{(0.397, 0.401)} & 0.215\tiny{(0.198, 0.235)} & 0.143\tiny{(0.132, 0.153)}\\
         & Gaussian$^{*}$ & 0.162\tiny{(0.161, 0.163)} & 0.399\tiny{(0.398, 0.401)} & 0.209\tiny{(0.193, 0.228)} & 0.154\tiny{(0.143, 0.165)}\\
         & IQN$^{*}$ & 0.161\tiny{(0.160, 0.162)} & 0.513\tiny{(0.510, 0.515)} & 0.210\tiny{(0.194, 0.229)} & 0.149\tiny{(0.139, 0.159)}\\
         & Informer & 0.324\ \tiny{(0.322, 0.327)} & 0.257\ \tiny{(0.255, 0.259)} & 0.257\ \tiny{(0.255, 0.259)} & 0.307\ \tiny{(0.305, 0.310)}\\
         & ETSformer & 0.267\ \tiny{(0.265, 0.270)} & 0.248\ \tiny{(0.246, 0.249)} & 0.249\ \tiny{(0.247, 0.251)} & 0.270\ \tiny{(0.268, 0.272)}\\
         & FEDformer & 0.249\ \tiny{(0.247, 0.251)} & 0.216\ \tiny{(0.214, 0.218)} & 0.205\ \tiny{(0.203, 0.207)} & 0.249\ \tiny{(0.247, 0.251)}\\
         & Autoformer & 0.235\ \tiny{(0.233, 0.237)} & 0.215\ \tiny{(0.213, 0.216)} & 0.205\ \tiny{(0.203, 0.206)} & 0.239\ \tiny{(0.237, 0.241)}\\
         & DLinear & 0.084\ \tiny{(0.083, 0.084)} & 0.277\ \tiny{(0.276, 0.278)} & \textbf{0.114\ \tiny{(0.107, 0.121)}} & 0.128\ \tiny{(0.121, 0.136)}\\
         & PatchTST & 0.080\ \tiny{(0.080, 0.081)} & 0.241\ \tiny{(0.239, 0.242)} & \textbf{0.114\ \tiny{(0.107, 0.121)}} & 0.094\ \tiny{(0.087, 0.101)}\\
         & TSMixer & 0.081\ \tiny{(0.081, 0.082)} & 0.241\ \tiny{(0.239, 0.242)} & \textbf{0.108\ \tiny{(0.102, 0.116)}} & 0.101\ \tiny{(0.095, 0.109)}\\
         & \method{}$^{*}$ & \textbf{0.075\ \tiny{(0.074, 0.075)}} & \textbf{0.201\ \tiny{(0.200, 0.203)}} & \textbf{0.113\ \tiny{(0.105, 0.121)}} & \textbf{0.068\ \tiny{(0.063, 0.074)}}\\
        \bottomrule
        \multicolumn{6}{l}{\footnotesize $^{*}$ Stochastic model}
    \end{tabular}
\end{table*}

\paragraph{Metrics and evaluation protocol}

We evaluate the models on a rolling-window basis. Namely, each dataset is divided into $W=100$ consecutive and overlapping time windows of equal length. For each time window, we train a separate model and evaluate it on $H$ time steps immediately following that window. This makes the evaluation more robust to outliers and allows us to assess the methods’ stability over time.  Furthermore, we can compute metrics on a more fine-grained scale and perform backtest. Importantly, this evaluation protocol is well-established in time series literature \citep{kupiec1995techniques,gonzalez2004forecasting, mcneil2015quantitative}, and differs from what is typically used for many transformer baseline models (only one evaluation window and no predictive power assessment, e.g., due to models' determinism). In what follows, we introduce the necessary definitions and describe how we aggregate the results. For additional information, see Appendix \ref{app:metrics}.

For error metrics, we use root-mean-square error (RMSE) and mean-absolute error (MAD),  well-established measures that can be computed for deterministic models. 
For stochastic models, we first compute the mean of the prediction and then compute the error, to achieve fairness in comparison with deterministic models. The metrics are given by the following formulas:
\begin{align*}
    &\text{RMSE}(w, s) = \\
    &\hspace{1.5cm} =\sqrt{ \frac{1}{H}\frac{1}{F^{s,w}} \sum_{h=1}^H \left(\frac{1}{I}\sum_{i=1}^{I} \hat{X}_{h,i}^{s,w} - X_{h}^{s,w}\right)^2}, \\
    &\text{MAD}(w, s) = \frac{1}{H}\frac{1}{F^{s, w}}\sum_{h=1}^H \left|\frac{1}{I}\sum_{i=1}^{I} \hat{X}_{h,i}^{s,w} - X_{h}^{s, w}\right|
\end{align*}
where $w$, $s$, $H$, and $I$ are the training window, series, horizon length, and number of samples, respectively. Furthermore, $X_h^{s, w}$ is the ground truth target value and $\hat{X}_{h, i}^{s, w}$ is the prediction of a given method, for series $s$, prediction horizon $h$, and simulation $i$. The normalizing factor  $F^{s,w}$ is the average absolute value of the ground truth series $s$ in the training window $w$. In this paper, we assume $H=24$ and $I=1024$.

For quantile metrics, we report a Quantile Loss (QL) and a Continuous Ranked Probability Score (CRPS) \citep{zamo2018estimation, berrisch2021crps}. Quantile Loss is defined as 
\[
    \text{QL}_{\alpha}(w, s) = \frac{2}{H}\frac{1}{F^{s,w}}\sum_{h=1}^H\left(\alpha - \mathbf{1}_{\Delta_{h, \alpha}^{s, w} \leq 0 } \right)\Delta_{h, \alpha}^{s, w},
\]

where $\Delta_{h, \alpha}^{s, w} = X_h^{s, w} - \hat{q}_{h}^{s,w}(\alpha) $ and $\hat{q}_{h}^{s,w}(\alpha)$ is the $\alpha$-quantile predicted by the model for series $s$, prediction horizon $h$, training window $w$, and $\alpha\in(0,1)$. To compute $\hat{q}_h^{s,w}(\alpha)$ we use an empirical estimator based on $I=1024$ samples from the model. CRPS summarizes the quantile loss over multiple quantile levels,
\[
    \text{CRPS}(w, s) = \frac{1}{M}\sum_{m=1}^M \text{QL}_{\alpha_m}(w, s),
\]
where $\alpha_m=\frac{m}{M+1}$ (we use $M=20$).

We aggregate error metrics (RMSE, MAD) and quantile metrics (QL, CRPS) for each dataset by summarizing the values over time series and windows via interquartile mean (IQM)~\citep{agarwal2021deep}. IQM discards bottom and top $25\%$ values and calculates the mean of the remaining $50\%$, making it robust to outliers. In our experiments, IQM gave results comparable to or more conservative than the median.

\paragraph{Backtest}

In addition to the above-mentioned metrics, we use the Kupiec backtest procedure \citep{kupiec1995techniques,mcneil2015quantitative} to measure the model's predictive performance. In particular, how well the model captures the data distribution and the corresponding quantiles. The test is based on the number of quantile violations, defined as:
\[
    \hat{v}_{h,\alpha}^s=\sum_{w=1}^{W}\mathbf{1}(\hat{q}_{h}^{s,w}(\alpha) < X_{h}^{s,w}).
\]
If the model correctly models the data randomness, i.e., the test's null hypothesis is true, $\hat{v}_{h,\alpha}^s$ has the Binomial distribution with parameters $(1-\alpha, W)$, see \citep[Section 9.3.1]{mcneil2015quantitative}.
The likelihood ratio test is given by 
\[
    T_{h,\alpha}^s = 2\log \frac{Bin(\hat{v}_{h,\alpha}^a, \hat{v}_{h,\alpha}^s/W, W)}{ Bin(\hat{v}_{h,\alpha}^s,1-\alpha, W)},
\]
where $Bin(k,p,n) = {n \choose k}p^k (1-q)^{n-k}$. Under the null hypothesis, $T_{h, \alpha}^s$ is asymptotically distributed as $\chi$-squared distribution with one degree of freedom, $\chi^2(1)$. Consequently, the $p$-value is given as $p\text{-value}_{h,\alpha}^s = \mathbb P(\chi^2(1) > T_{h, \alpha}^s)$. 

In experiments, we report the fraction of time series and horizons for which their $p$-value in the underlying likelihood ratio test is at least $\gamma\in(0,1)$: 
\[
    p\text{-value}_{\alpha} = \frac{1}{SH}\sum_{h=1}^H\sum_{s=1}^S\mathbf{1}(p\text{-value}_{h, \alpha}^s\ge \gamma).
\]
Informally, this tells us for how many time series and horizons the model correctly predicts tail events according to the Kupiec test with significance level $\gamma$; here $\gamma=5\%$. 

\begin{table*}[t]
    \centering
    \caption{Performance of the models in terms of quantile metrics QL and CRPS (lower values are better). Each entry is computed using IQM with $90\%$ bootstrap confidence intervals (in brackets) computed over $W=100$ training windows and all time series in each dataset.} 
    \label{tab:stochastic_preformance}

    \begin{tabular}{llllll}
        \toprule
        metric & model &           \electricity{} &            \traffic{} &                \ettm{} &            \weather{} \\
        \midrule
        CRPS & t-student & 0.11\ \tiny{(0.105, 0.106)} & 0.24\ \tiny{(0.235, 0.237)} & 0.18\ \tiny{(0.166, 0.198)} & 0.12\ \tiny{(0.108, 0.125)}\\
             & Laplace & 0.11\ \tiny{(0.106, 0.107)} & 0.24\ \tiny{(0.234, 0.236)} & 0.18\ \tiny{(0.166, 0.199)} & 0.12\ \tiny{(0.107, 0.124)}\\
             & Gaussian & 0.1\ \tiny{(0.103, 0.104)} & 0.23\ \tiny{(0.229, 0.23)} & 0.17\ \tiny{(0.157, 0.186)} & 0.12\ \tiny{(0.114, 0.131)}\\
             & IQN & 0.1\ \tiny{(0.101, 0.102)} & 0.27\ \tiny{(0.271, 0.273)} & 0.17\ \tiny{(0.155, 0.185)} & 0.11\ \tiny{(0.103, 0.118)}\\
             & \method{} & \textbf{0.04\ \tiny{(0.042, 0.042)}} & \textbf{0.08\ \tiny{(0.081, 0.082)}} & \textbf{0.07\ \tiny{(0.068, 0.078)}} & \textbf{0.04\ \tiny{(0.037, 0.044)}}\\
        \midrule
        $\text{QL}_{50\%}$ & t-student & 0.13\ \tiny{(0.125, 0.126)} & 0.27\ \tiny{(0.272, 0.274)} & 0.19\ \tiny{(0.172, 0.204)} & 0.12\ \tiny{(0.116, 0.134)}\\
         & Laplace & 0.13\ \tiny{(0.126, 0.127)} & 0.27\ \tiny{(0.272, 0.273)} & 0.19\ \tiny{(0.171, 0.204)} & 0.12\ \tiny{(0.115, 0.134)}\\
         & Gaussian & 0.12\ \tiny{(0.124, 0.125)} & 0.28\ \tiny{(0.274, 0.276)} & 0.18\ \tiny{(0.168, 0.199)} & 0.14\ \tiny{(0.126, 0.145)}\\
         & IQN & 0.12\ \tiny{(0.123, 0.125)} & 0.34\ \tiny{(0.337, 0.34)} & 0.18\ \tiny{(0.168, 0.2)} & 0.13\ \tiny{(0.119, 0.136)}\\
         & \method{} & \textbf{0.06\ \tiny{(0.055, 0.056)}} & \textbf{0.1\ \tiny{(0.103, 0.104)}} & \textbf{0.09\ \tiny{(0.086, 0.098)}} & \textbf{0.05\ \tiny{(0.047, 0.056)}}\\
        \midrule
        $\text{QL}_{75\%}$ & t-student & 0.11\ \tiny{(0.11, 0.111)} & 0.25\ \tiny{(0.252, 0.254)} & 0.17\ \tiny{(0.153, 0.182)} & 0.11\ \tiny{(0.102, 0.12)}\\
         & Laplace & 0.11\ \tiny{(0.111, 0.112)} & 0.25\ \tiny{(0.253, 0.255)} & 0.17\ \tiny{(0.153, 0.183)} & 0.11\ \tiny{(0.101, 0.12)}\\
         & Gaussian & 0.11\ \tiny{(0.107, 0.108)} & 0.25\ \tiny{(0.247, 0.249)} & 0.16\ \tiny{(0.146, 0.173)} & 0.13\ \tiny{(0.118, 0.137)}\\
         & IQN & 0.11\ \tiny{(0.108, 0.109)} & 0.32\ \tiny{(0.317, 0.32)} & 0.16\ \tiny{(0.143, 0.17)} & 0.11\ \tiny{(0.102, 0.118)}\\
         & \method{} & \textbf{0.05\ \tiny{(0.046, 0.047)}} & \textbf{0.09\ \tiny{(0.09, 0.091)}} & \textbf{0.08\ \tiny{(0.072, 0.082)}} & \textbf{0.04\ \tiny{(0.04, 0.047)}}\\
        \midrule
        $\text{QL}_{95\%}$ & t-student & 0.06\ \tiny{(0.06, 0.061)} & 0.16\ \tiny{(0.162, 0.164)} & 0.12\ \tiny{(0.103, 0.129)} & 0.06\ \tiny{(0.053, 0.068)}\\
         & Laplace & 0.06\ \tiny{(0.057, 0.058)} & 0.16\ \tiny{(0.155, 0.157)} & 0.11\ \tiny{(0.102, 0.128)} & 0.06\ \tiny{(0.05, 0.064)}\\
         & Gaussian & 0.06\ \tiny{(0.056, 0.057)} & 0.14\ \tiny{(0.139, 0.14)} & 0.09\ \tiny{(0.082, 0.105)} & 0.06\ \tiny{(0.051, 0.063)}\\
         & IQN & 0.05\ \tiny{(0.052, 0.053)} & 0.15\ \tiny{(0.152, 0.153)} & 0.09\ \tiny{(0.076, 0.098)} & 0.05\ \tiny{(0.045, 0.055)}\\
         & \method{} & \textbf{0.02\ \tiny{(0.016, 0.017)}} & \textbf{0.04\ \tiny{(0.037, 0.037)}} & \textbf{0.03\ \tiny{(0.029, 0.035)}} & \textbf{0.02\ \tiny{(0.015, 0.017)}}\\
        \bottomrule
    \end{tabular}
\end{table*}

\section{Experiments}\label{sec:main_results}

\looseness=-1 
The key finding of this section is that \method{} surpasses existing state-of-the-art methods in RMSE and MAD metrics, as detailed in Section \ref{sec:main_result_rmse_mad}. This highlights the advantage of stochastic models, showing they not only apply to a broader array of scenarios but also provide more accurate pointwise predictions compared to deterministic models.
We further analyze the behavior of these models from a distributional perspective in Sections~\ref{sec:analysis_ql_crps} to \ref{sec:analysis_backtest}. It is important to note that such \emph{stochastic analysis is not possible for most of the contemporary baselines due to their determinism}. Our model outperforms stochastic baselines on QL and CRPS (Section \ref{sec:analysis_ql_crps}) and does a good job at predicting quantiles (Section \ref{sec:analysis_backtest}) for datasets \electricity{} and \traffic{}. However, all models struggle on \ettm{} and \weather{} datasets, highlighting the need for rigorous evaluation and an inflow of new ideas to the field. Finally, we confirm in Section \ref{sec:permutation} that \method{} does not suffer from weaknesses pointed out by \citet{Zeng2022AreTE}. 
Additional experiments are available in Appendix: for ablations on scaling, normalization, $\beta$, discretization, or embeddings, see Appendix~\ref{app:ablations}; for an application of time series forecasting to an atypical task of predicting hand-drawn images, see Appendix~\ref{app:hand-drawn-images}.

\subsection{Main result: performance on error metrics} \label{sec:main_result_rmse_mad}

Our experiments reveal that \method{} outperforms baselines on \electricity{}, \traffic{}, and \weather{} datasets, see Table \ref{tab:deterministic_preformance}. 
Importantly, \method{}'s confidence intervals do not overlap with those of the other models. On \ettm{}, \method{} matches the performance of DLinear, PatchTST, and TSMixer, as their confidence intervals overlap. 

Interestingly, stochastic methods that employ parametric marginal distributions (Gaussian, Laplace, and t-student) often match or surpass the performance of more intricate transformer-based models, e.g., Informer, ETSformer, FEDformer, Autoformer on \electricity{}. This observation, especially when combined with further improvement to those results when applying \method{}, highlights the value of stochastic modeling in time series forecasting. 

We hypothesize that the reasons for the overall \method{}'s strong performance are threefold. First, the stochastic nature of \method{} allows us to simulate multiple predictions and compute their mean achieving a variance reduction effect. Second, our model benefits from the transformer's dense learning signal in the form of the next token prediction objective. In particular, this incentivizes the emergence of rich internal data representations. Third, \method{} models complex non-parametric marginal distributions allowing to capture stochastic behavior of individual time series. This is possible due to our tokenization strategy, which also makes it easy for the transformer to handle real-valued data.

\subsection{Analysis: quantile metrics}\label{sec:analysis_ql_crps}

In this section, we investigate the models' performance on quantile metrics QL and CRPS\footnotemark{}, as defined in Section \ref{sec:experiments}. These metrics are useful in ranking the models and the results are presented in Table \ref{tab:stochastic_preformance}. 
\footnotetext{Such an analysis is unavailable for deterministic baselines considered in this paper, i.e., Informer,  ETSformer,  FEDformer,  Autoformer,  DLinear,  PatchTST, and TSMixer.} 
We observe that \method{} outperforms other methods on CRPS by a significant margin. This is encouraging, as this metric summarizes how the model fares in predicting quantiles at multiple levels. The results for QL$_{50\%}$, QL$_{75\%}$, and QL$_{95\%}$ confirm this conclusion. Having ranked the models, we conduct a detailed analysis of their performance in the next section.

\begin{table*}[t]
    \centering
    \caption{Fraction of horizon-series pairs in each dataset with $p$-values exceeding $5\%$ for Kupiec's PoF test on each quantile.}
    \label{tab:p_values_per_quantile}
    \begin{tabular}{llllll}
        \toprule
        metric &                      model & \electricity{} & \traffic{} & \ettm{} & \weather{} \\
        \midrule
        $p\text{-value}_{50\%}$ & t-student & 72.7\% & \textbf{89.9\%} & 85.1\% & 68.5\%\\
                                & Laplace   & 73.7\% & 89.5\% & 88.1\% & 65.1\%\\
                                & Gaussian  & 73.2\% & 88.6\% & 90.5\% & \textbf{70.4\%} \\
                                & IQN       & 74.0\% & 46.3\% & \textbf{94.0\%} & 65.5\%\\
                                & \method{} & \textbf{88.0\%} & 88.9\% & 63.1\% & 48.8\%\\
        \midrule
        $p\text{-value}_{75\%}$ & t-student & 24.6\% & 6.7\% & 0.0\% & 15.3\%\\
                                & Laplace   & 22.7\% & 6.6\% & 0.0\% & 14.3\%\\
                                & Gaussian  & 36.4\% & 21.1\% & 5.4\% & 17.7\%\\
                                & IQN       & 24.1\% & 17.0\% & 1.8\% & 9.7\%\\
                                & \method{} & \textbf{78.0\%} & \textbf{80.7\%} & \textbf{11.9\%} & \textbf{40.3\%}\\
        \midrule
        $p\text{-value}_{95\%}$ & t-student & 0.7\% & 0.0\% & 0.0\% & 11.9\%\\
                                & Laplace   & 1.6\% & 0.2\% & 0.0\% & 13.7\%\\
                                & Gaussian  & 1.6\% & 0.7\% & 0.0\% & 17.1\%\\
                                & IQN       & 3.0\% & 18.8\% & 0.0\% & 3.4\%\\
                                & \method{} & \textbf{67.2\%} & \textbf{71.8\%} & 0.0\% & \textbf{21.2\%}\\
        \bottomrule
    \end{tabular}
\end{table*}

\subsection{Analysis: backtest}\label{sec:analysis_backtest}

In this section, we study the methods' ability to model the data distribution. The findings are presented in Table~\ref{tab:p_values_per_quantile}, which reports the outcomes of a backtesting procedure proposed by \citet{kupiec1995techniques} and described in Section~\ref{sec:experiments} and Appendix~\ref{app:metrics}. Namely, the table's entries represent a fraction of all time series and horizons for which the models passed the test. This number summarizes how accurately each model captures the data distribution.

Table \ref{tab:p_values_per_quantile} indicates that \method{} achieves the highest performance at quantile levels $75\%$ and $95\%$ across all datasets. For median, 
the results vary: \method{} leads on \electricity{}, matches the performance of t-student, Laplace, and Gaussian on \traffic{}, and performs adequately on \weather{}, but does not surpass the baselines.

The results presented in Table \ref{tab:p_values_per_quantile} reveal two key insights. First, there is clearly a negative correlation between $\alpha$ and models' ability to predict $\alpha$-quantile. For many models this drop in performance is severe. Yet, \method{} stands out as it consistently maintains good performance across all quantile levels on \electricity{} and \traffic{}. Second, there is a stark difference in the difficulty of data distribution modeling between datasets. This is particularly visible for \ettm{} and \weather{}, where all the methods achieve good or very good results for modeling the median but collapse for higher quantile levels. We suspect this difficulty stems from the high volatility and unpredictability of time series in these specific datasets.

We notice that the Laplace and t-student approaches do not perform particularly well at $75\%$ and $95\%$ quantile levels on any dataset. This is surprising, given that these distributions are characterized by heavy tails, making them natural candidates for solving the task at hand. 

\begin{figure*}[t]
    \centering
    \includegraphics[width=\textwidth]{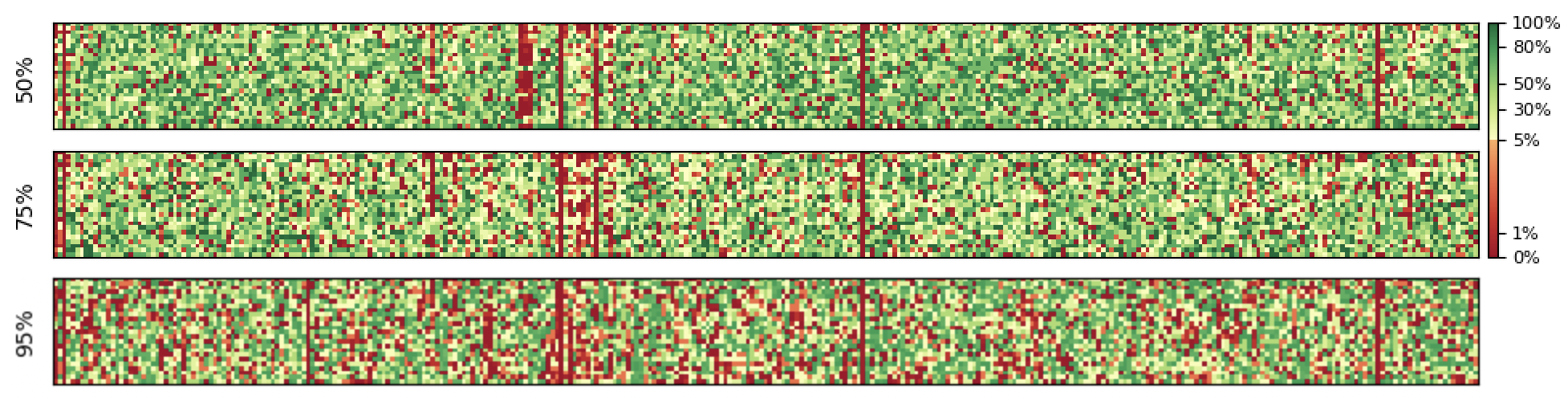}
    \caption{Color-coded p-values from backtesting \method{} on \electricity{}. Each rectangle represents the results for backtesting \method{} on one of the following levels: $50\%$, $75\%$, and $95\%$. The height of a rectangle equals $H=24$, corresponding to the number of prediction steps, and a width equals to $S=321$, the number of time series in \electricity{}. 
    }
    \label{fig:p_values_strips}
\end{figure*}

Finally, it is instructive to delve into \method{}'s modeling capabilities on a dataset where it excels. Figure \ref{fig:p_values_strips} showcases p-values from backtesting \method{} on \electricity{} across three quantile levels ($50\%, 75\%$ and $95\%$), 24 prediction horizons, and 321 individual time series. Each cell in the depicted matrices corresponds to a single p-value, color-coded green for acceptance or red for rejection at the $5\%$ significance level. Green dominates all matrices, indicating a generally high model performance. However, we observe that the amount of red color increases with higher quantile levels, reflecting the rates from Table \ref{tab:p_values_per_quantile}: $88.0\%$ for $\alpha=50\%$, $78.0\%$ for $\alpha=75\%$, and $67.2\%$ for $\alpha=95\%$. Additionally, we can see that for certain time series, predicting quantiles across any horizon remains particularly challenging, as indicated by the vertical red lines.

\begin{table*}[t]
    \centering
    \caption{Performance of \method{} after a random shuffling of input during inference.}
    \label{tab:shuffling}
    \begin{tabular}{lllllll}
        \toprule
        metric & model               & \electricity{}             & \traffic{}                 & \ettm{}                    & \weather{}                \\
        \midrule
        MAD  & \method{} + shuffling & 0.283\tiny{(0.282, 0.285)} & 0.590\tiny{(0.589, 0.592)} & 1.306\tiny{(1.199, 1.426)} & 0.210\tiny{(0.197, 0.225)}\\
             & \method{}             & 0.057\tiny{(0.057, 0.058)} & 0.114\tiny{(0.114, 0.115)} & 0.092\tiny{(0.086, 0.098)} & 0.056\tiny{(0.052, 0.061)}\\
        
        \midrule
        RMSE & \method{} + shuffling & 0.328\tiny{(0.327, 0.329)} & 0.733\tiny{(0.731, 0.735)} & 1.398\tiny{(1.291, 1.517)} & 0.239\tiny{(0.224, 0.255)}\\
             & \method{}             & 0.075\tiny{(0.074, 0.075)} & 0.201\tiny{(0.200, 0.203)} & 0.113\tiny{(0.105, 0.121)} & 0.068\tiny{(0.063, 0.074)}\\
        \bottomrule
    \end{tabular}
\end{table*}

\subsection{Analysis: input permutation}\label{sec:permutation}

Transformer models, including Informer, Autoformer, and FEDformer have been criticized for insufficient handling of temporal dependencies, and some even exhibit permutational invariance~\citep{Zeng2022AreTE}. This can be viewed as a serious issue, as the sequential order often plays a crucial role in time series modeling. Additionally, \citet{Zeng2022AreTE} demonstrated that these methods underperform with respect to a simple MLP-based model, DLinear. We show that \method{} suffers a significant drop in performance when randomly shuffling the input (Table~\ref{tab:shuffling}) and that its performance is superior to DLinear (Table~\ref{tab:deterministic_preformance}), addressing the main concerns of \citep{Zeng2022AreTE} about using transformers for time series modeling.

\section{Limitations and future work}\label{sec:limitations}

\paragraph{Quadratic memory complexity} 
This issue limits our options for making long-horizon predictions. It is further exacerbated by linear scaling with precision. In future work, we will explore various computational and memory efficiency improvements \citep{tay2022efficient}. 
\looseness-100

\paragraph{Backtesting}

Backtesting is an effective tool to measure the predictive performance of the model, particularly in situations when adequate risk capturing is of special importance. For large transformer models, this can be computationally expensive. However, a low-rank adaptation technique~\citep{hu2021lora} holds some promise in alleviating this problem.

\paragraph{Probability distribution function}

Generic transformer models allow sampling trajectories but do not give easy access to the probability distribution functions. 
It would be interesting to see how this obstacle can be overcome.

\paragraph{Uncertainty-aware model}

It would be beneficial to study whether \method{} can evaluate its own uncertainty, similarly to \citep{kadavath2022language}.

\paragraph{Normalization}

Normalization has some non-trivial interactions with digit distribution, as additive fluctuations do not lead to the Newcomb-Benford law \citep{hill1995significant}.
In future work, we would like to test other recently developed normalization schemes \citep{kim2022reversible,nie2022time}.

\paragraph{Explainability}

Reverse engineering transformer models remains a challenging task. Nevertheless, recent advancements in the so-called mechanistic interpretability, see \citep{nanda2023progress,chughtai2023toy, wang2023interpretability}, offer promising approaches to address this issue. Investigating the underlying circuits of learned time series transformers is an exciting avenue for future research.

\section{Conclusions}

In this paper, we propose \method{}, a general-purpose transformer stochastic time series model. We focus on evaluating the model’s predictive capabilities, employing a rolling window analysis, a well-known time-series procedure. We show that \method{} outperforms the current state-of-the-art models on MAD and RMSE on four popular datasets: \electricity{}, \traffic{}, \ettm{}, \weather{}. Similarly, we demonstrate that \method{} surpasses its stochastic peers on QL and CRPS. We complement these results with a thorough analysis using the Kupiec backtest. We show how the models behave on datasets of varying complexity and at different quantile levels. We go as deep as showing the test results for each prediction period, quantile level, and time series in \electricity{}. We close the analysis showing that \method{} is not permutationally invariant. This addresses a prominent critique directed towards some contemporary time series transformer models, raised by \citep{Zeng2022AreTE}. Appendix contains results of additional ablation studies.

\section*{Impact Statement}

This paper presents work whose goal is to advance the field of Machine Learning in general, and time series modeling in particular. There are many potential societal consequences of our work, none of which we feel must be specifically highlighted here.

\section*{Acknowledgments}

We gratefully acknowledge Polish high-performance computing infrastructure PLGrid (HPC Centers: ACK Cyfronet AGH, CI TASK) for providing computer facilities and support within computational grant numbers PLG/2023/016264. Piotr Miłoś was supported by National Science Center Poland under the grant agreement 2019/35/O/ST6/03464.

\bibliography{bibliography}
\bibliographystyle{icml2024}

\newpage
\appendix
\onecolumn


\section{Model's architecture}\label{app:architecture}

\subsection{Transformer }\label{app:transformer}

The transformer \citep{vaswani2017attention} is a deep neural network model that relies on the attention mechanism. We employ a decoder-only variant of this architecture, a common choice in natural language modeling \citep{radford2019language, brown2020language}. A transformer model takes a sequence of tokens as input and outputs a sequence of logits over all the tokens. The transformer's central object is a residual stream, which serves as a working memory for the model. Each token undergoes the initial embedding into a space of dimension $d_\text{model}$, after which it is processed by $N$ decoder layers. These layers consist of attention and feed-forward (FF) blocks. These blocks can be conceptualized as operations on the residual stream. Specifically, each one of $n_\text{heads}$ heads inside the attention block operates on dimension $d_\text{head}=d_\text{model}/n_\text{heads}$ and moves information between positions, while the FF block writes or removes from tokens' residual streams \citep{nanda2023progress}. The FF block incorporates nonlinearity (we use the Gaussian Error Linear Unit or GELU) and is composed of a two-layer multilayer perceptron (MLP) with a hidden space of dimension $d_\text{ff}$. The inputs to attention, feed-forward  blocks, and the output logits layer are processed by layer norms.

The causal self-attention mechanism for each element of the sequence assigns different weights to each preceding token based on its relevance to the current element, which allows for flexible and context-aware representations. Independent attention operations are executed $H$ times, once for every attention head, enabling the model to learn distinct representations. For each head, the attention function can be expressed as:

\[
    \text{Attention}(X) = \text{softmax}\left(\frac{QK^T + M}{\sqrt{d_{\text{head}}}}\right)V,
\]

where $Q = XW^Q$, $K = XW^K$ and $V = XW^V$ for $W^Q$, $W^K$, $W^V \in \mathbb{R}^{d_\text{model} \times d_\text{head}}$ are trainable linear projections of the residual stream normalized by layer norm, and $M$ is a causal mask with zeros below the diagonal and negative infinity elsewhere. The outputs of attention heads are concatenated and linearly projected to $d_\text{model}$. We also extend the attention mechanism by incorporating rotary encodings \citep{su2021roformer}. For more details on the transformer architecture, please refer to the original work of \citet{vaswani2017attention}.

\subsection{Normalization }\label{app:normalization}

Inspired by \citep{salinas2020deepar}, we scale model input $X$ by dividing it by $\mu_X = r + \frac{1}{T} \sum_{t=1}^{T} |X_t|$, where $T$ denotes the sequence length and $r$ is a positive constant. We invert the scaling at the model's output. During prediction, we calculate the scaling factor based only on the context to prevent information leaks from the prediction horizon.

\subsection{Digit Tokenization}\label{app:digitfactorization}

\begin{algorithm}
    \caption{Digit Factorization}\label{alg:digit_extraction}
    \begin{tabular}{ l c l }
        \textbf{Requires: }
        & $x$ & input to be discretized into $p$ digits in base $v$ \\
        & $v$ & vocabulary size, which acts as the number base \\
        & $p$ & precision to be used in the discretization \\
        & $l, h$ & lower and upper bounds used for squashing $x$ into $[0,1]$ 
    \end{tabular}
    \begin{algorithmic}[1]
        \Function{DigitFactorization}{$x$; $p$, $v$, $l$, $h$}
            \State $x \gets \text{clip}((x - l) / (h - l), 0, 1)$
            \Comment{Squash to $[0,1]$}
            \State $D \gets$ []
            \State $r \gets 1$
            \For{$i = 1$ to $p$}
                \State $r \gets r / v$
                \State $d \gets \text{min} ( \lfloor x / r\rfloor , v - 1 )$
                \State $D.\text{append}(d)$
                \State $x \gets x - d \cdot r$
            \EndFor
            \State \Return $D$
        \EndFunction
    \end{algorithmic}
\end{algorithm}

The input to the transformer decoder (Section \ref{app:transformer}) is normalized (see Section \ref{app:normalization}) and discretized into digits, see Algorithm \ref{alg:digit_extraction}. The selected number base $v$ equals the vocabulary size used for the model. We squash the input $x$ into  $[0,1]$ interval and extract $p$ most significant digits. Hence, a sequence of length $T$ is discretized into a sequence of $p\cdot T$ digits. We notice that the higher the precision, the longer the input. Put differently, with a fixed context length, increasing $p$ reduces the amount of raw data fed into the model. This is a limitation caused by the quadratic complexity of transformers, see also Section \ref{sec:limitations}.

\section{Hyperparameters}\label{app:hyperparameters}

\subsection{Hyperparameters of baselines}

The hyperparameters used to configure and train the baselines are given in Table~\ref{tab:hyperparams_general} and Table~\ref{tab:params_per_dataset}. The parameters were chosen according to configurations supplied in the respective public repositories, except for \texttt{pred\_len}, \texttt{seq\_len} and \texttt{label\_len}, which were adjusted to match the training and evaluation setup of this work. 

\begin{table*}[htpb]
    \centering
    \caption{
        Hyperparameters of baselines. Names refer to parameters in the corresponding code repositories.
    }
    \begin{tabular}{rccccccc}
                              &      Autoformer &       ETSformer &       FEDformer &  Informer &   PatchTST &     DLinear &     TSMixer \\
        \midrule
        d\_model              &             512 &             512 &             512 &       512 &        128 &         --- &          64 \\
        d\_ff                 &            2048 &            2048 &            2048 &      2048 &        256 &         --- &       32-64 \\
        activation            &            gelu &            gelu &            gelu &      gelu &       gelu &        gelu &        relu \\
        e\_layers / n\_block  &               2 &               2 &               2 &         2 &          3 &         --- &       2 - 8 \\
        d\_layers             &               1 &               1 &               1 &         1 &        --- &         --- &         --- \\
        n\_heads              &               8 &               8 &               8 &         8 &         16 &         --- &         --- \\
        seq\_len              &             232 &             232 &             232 &       232 &        232 &         232 &         232 \\
        pred\_len             &              24 &              24 &              24 &        24 &         24 &          24 &          24 \\
        label\_len            &             116 &               0 &             116 &       116 &         48 &         48  &         --- \\
        distilling in encoder &            True &            True &            True &      True &       True &         --- &         --- \\
        train\_epochs         &            3-10 &            3-10 &            3-10 &      3-10 &        100 &         100 &         100 \\
        batch\_size           &              32 &              32 &              32 &        32 &   24 - 128 &     16 - 32 &          32 \\
        learning\_rate        &            1e-4 &     1e-5 - 1e-3 &            1e-4 &      1e-4 &       1e-4 & 1e-4 - 5e-2 & 1e-4 - 1e-3 \\
        dropout               &            0.05 &            0.05 &            0.05 &      0.05 &        0.2 &         --- &   0.3 - 0.9 \\
        loss                  &             mse &             mse &             mse &       mse &        mse &         mse &         mse \\
        lradj                 &           type1 &           type1 &           type1 &     type1 & TST, type3 &       type3 &         --- \\
        ES patience           &               3 &               3 &               3 &        3  &    10 - 20 &         --- &           5 \\
        RevIn                 &            ---  &            ---  &             --- &       --- &       True &         --- &        True \\
        \bottomrule
        \label{tab:hyperparams_general}
    \end{tabular}
\end{table*}

\begin{table*}[htbp]
    \centering
    \caption{Hyperparameters of baseline models which are adjusted per-dataset.}
    \begin{tabular}{llcccc}
        model &  parameter &  \electricity{} & \traffic{} & \ettm{} & \weather{} \\
        \midrule
        Autoformer & train\_epochs & 10 & 3 & 10 & 10 \\
        ETSformer  & train\_epochs & 10 & 3 & 10 & 10 \\
                   & K & 3 & 3 & 3& 1 \\
                   & learning\_rate & 3e-4 & 1e-3 & 1e-5 & 1e-3 \\
        FEDformer  &  train\_epochs & 10 & 3 & 10 & 10 \\
        Informer   &  train\_epochs & 10 & 3 & 10 & 10 \\
        PatchTST   & early stopping patience & 10 & 10 & 20  & 20 \\
                   & batch\_size & 32 & 24 & 128  & 128 \\ 
        DLinear    & learning\_rate & 1e-3 & 5e-2 & 1e-4 & 1e-3 \\
                   & batch\_size & 16 & 16 & 16 & 32 \\
        TSMixer    & learning\_rate & 1e-4 & 1e-4 & 1e-3 & 1e-4 \\
                   & n\_block & 4 & 8 & 2 & 4 \\
                   & d\_ff & 64 & 64 & 64 & 32 \\
                   & dropout & 0.7 & 0.7 & 0.9 & 0.3 \\
        \bottomrule
    \end{tabular}
    \label{tab:params_per_dataset}
\end{table*}

\newpage
\subsection{Hyperparameters of \method{}}\label{sec:tsgt_params}

We performed a grid search over the number of layers $(4, 6, 8)$, multifactor constant which controls the learning rate ($0.01, 0.03, 0.1$), the number of attention heads ($2, 4, 8$), dimension of model ($64, 128, 256, 512$), dropout ($0.1, 0.2, 0.3, 0.4, 0.5$), precision ($1, 2, 3$), and vocabulary size ($1024, 512, 10$). Based on this grid search, we chose a universal set of hyperparameters which is summerized in Table \ref{tab:hyperparams_model}.

\begin{table*}[htbp]
    \centering
    \caption{Hyperparameters for \method{}}
    \begin{tabular}{rl}
        \toprule
        $d_\text{model}$     &           256    \\
        $d_\text{ff}$        &           512    \\
        activation           &           gelu   \\
        $N$                  &           6      \\
        $n_\text{heads}$     &           4      \\
        $T$                  &           256    \\
        $H$                  &           24     \\
        train\_steps         &           $10^5$ \\
        batch\_size          &           16     \\
        multifactor constant &           0.03   \\
        dropout              &           0.1    \\
        precision            &           3      \\
        vocab\_size          &           10     \\
        \bottomrule
    \end{tabular}
    \label{tab:hyperparams_model}
\end{table*}

\section{Stochastic baselines}\label{app:stochastic_baselines}

\subsection{Distribution-based baselines}

As a stochastic baseline for our model, inspired by \citet{li2019enhancing}, we implement models that predict the parameters of the probability distributions. We use a transformer backbone and train the models to maximize the log probability of training data. We compare against 3 different symmetric distributions: t-student, Laplace, and Gaussian. For all those, we train the scale and location parameters of the distribution. We use training pipeline based on next token prediction. The hyperparameters of the model and the optimizer are fixed. They are kept the same as those used for \method{} (see Table~\ref{tab:hyperparams_model}). 

Similar to \citet{li2019enhancing} we use convolutional attention, and set the kernel width to 3, after running a hyperparameter search. We also use additional covariates. All models use series index, a logarithm of the timestep, and an hour, as well as either a day in case of \electricity{} and \traffic{} datasets, or a minute in case of \ettm{} and \weather{}.

\subsection{Implicit Quantile Baseline}

Another stochastic baseline is inspired by \citet{gouttes2021probabilistic}. We train a transformer architecture with Implicit Quantile Module (IQM) that learns the inverse of the cumulative probability function using Quantile Loss. Again, the hyperparameters of the Transformer backbone and training hyperparameters are fixed. They are kept the same as those used to tune \method{} (see Table~\ref{tab:hyperparams_model}). The only hyperparameter left is the width of the embedding in the IQM head, which is kept at 256 to match the size of the hidden representation of the transformer. We train a transformer with IQM head to perform next token prediction, by minimizing quantile loss. For more details on the IQM and its exemplar application for time series modeling please refer to \citet{gouttes2021probabilistic}.

\section{Deterministic baselines }\label{app:baselines}

Among others, we used the following methods as baselines: FEDformer \citep{zhou2022fedformer},  Informer \citep{zhou2021informer},  Autoformer \citep{wu2021autoformer}, ETSformer \citep{woo2022etsformer}.

All the above-mentioned baseline models, except for ETSformer, include additional data to the input; these are composed of timestamps (e.g., minute, hour, week, month, and year) or other types of information (e.g., holidays and events).  This is done by adding additional learnable encoding to the architecture. The baselines deterministically output a prediction trajectory. 

Each model proposes some efficiency improvements to reduce the quadratic memory complexity of the transformer: \citep{li2019enhancing} proposes log-sparse attention while 
\citep{zhou2021informer,zhou2022fedformer,wu2021autoformer} use low-rank property of attention.

Furthermore, each baseline implements a time series-specific architectural change: \citep{li2019enhancing} introduces $1$d-conv attention, \citep{zhou2021informer} designs a decoder to produce prediction trajectories directly, \citep{wu2021autoformer} introduces a seasonal-trend decomposition with an auto-correlation block, \citep{zhou2022fedformer, woo2022etsformer} explore attention mechanisms in the frequency domain.

\section{Data}\label{app:data}

\begin{figure}
    \centering
    \includegraphics[width=0.99\textwidth]{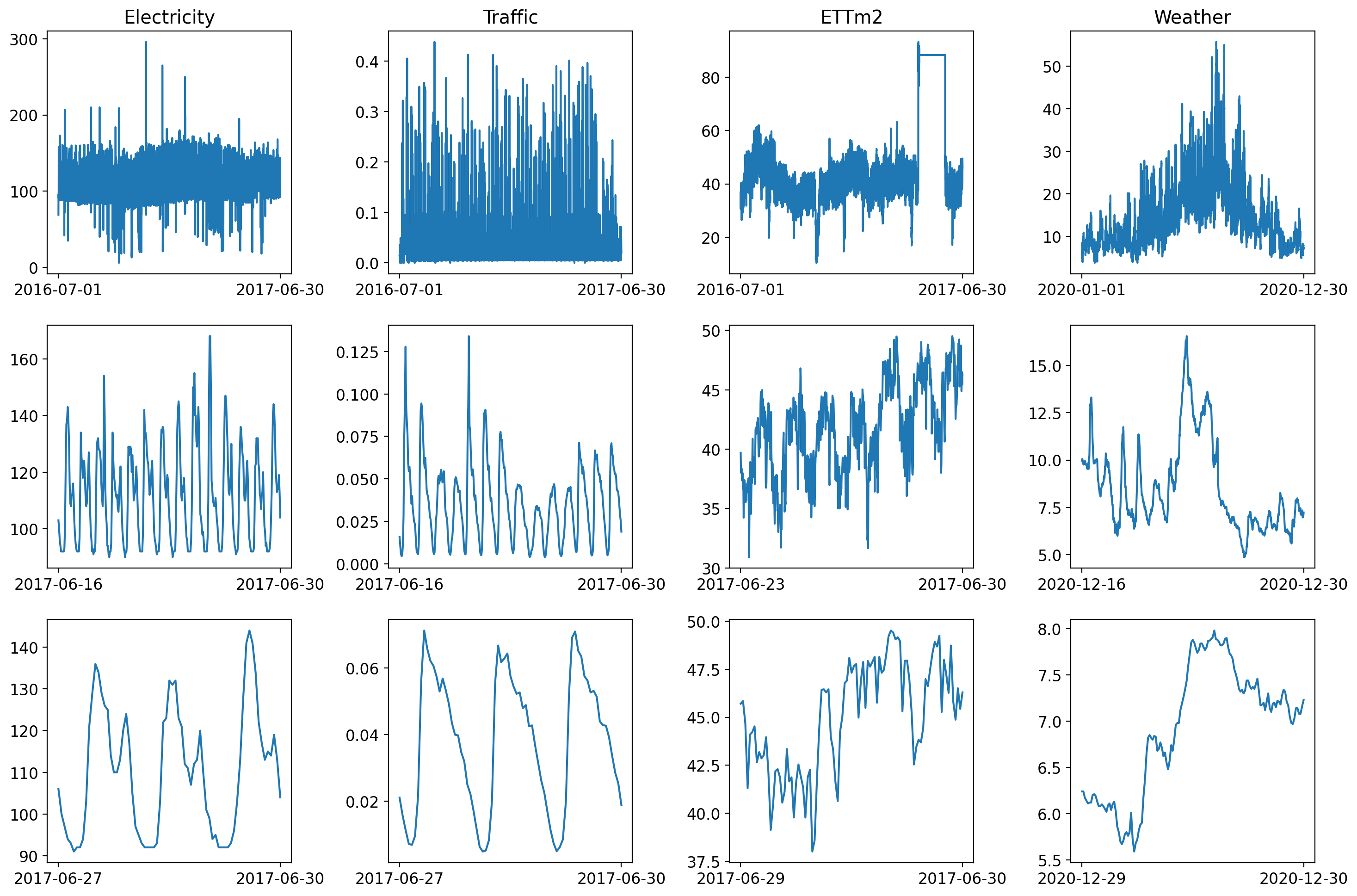}
    \caption{Visualization of data from each of the four real-life datasets. The first row displays a year's worth of data for a selected series. The second and the third rows zoom in on the suffix of those trajectories.}
    \label{app:fig:data}
\end{figure}

We use four real-life datasets popular in the field: \electricity{}, \traffic{}, \ettm{}, and \weather{}, see \cite{wu2021autoformer} and the references therein. Exemplary trajectories are visualized in Figure~\ref{app:fig:data}, and datasets statistics are provided in Table~\ref{tab:app:data}. 

The \electricity{} dataset records the hourly electricity consumption of 321 customers from July 2016 to July 2019. The \traffic{} dataset comprises hourly Caltrans data from July 2016 to July 2018 on road occupancy rates for San Francisco Bay area freeways, measured by $862$ sensors. Both \electricity{} and \traffic{} are characterized by a large number of periodic patterns. 

The \ettm{} dataset was collected from electricity transformers in a province of China and contains seven time series representing, e.g., load and oil temperature recorded every 15 minutes between July 2016 and July 2018. It includes negative values, outliers, and erratic behavior. 

The \weather{} dataset holds the recordings of $21$ meteorological indicators collected every $10$ minutes between January 2020 and January 2021 in Germany. The data includes air temperature or humidity and is characterized by varying scales, outliers, and a significant number of zero values.

\begin{table*}[]
    \centering
    \begin{tabular}{c|c|c|c}
         Dataset & Total length & Number of sequences & Frequency  \\
         \midrule
         \electricity{} & 26,304 & 321 & 1H  \\
         \traffic{}     & 17,544 & 862 & 1H  \\
         \ettm{}        & 69,680 &   7 & 15T \\
         \weather{}     & 52,704 &  21 & 10T \\
         \bottomrule
    \end{tabular}
    \caption{Statistics of the datasets used in this paper.}
    \label{tab:app:data}
\end{table*}

\section{Metrics}\label{app:metrics}

\paragraph{Error metrics} 

Error metrics are defined in the paragraph `Metrics and evaluation protocol' in Section \ref{sec:experiments}. The normalizing constant $F^{s,w}$, for $w\in\{1,\ldots, W\}$ and $s\in\{1, \ldots, S\}$ is defined as $F^{s,w} = \frac{1}{T_{end}^w - T_{start}^w} \sum_{k=T_{start}^w}^{T_{end}^w} |X_k^{s,w}|$, where $T_{start}^w,T_{end}^w$ are the training window's $w$ beginning and the end, and $X^{s,w}$ is the ground truth series.

\paragraph{Backtest metrics} 

We use the Kupiec proportion of failures (POF) likelihood ratio test \citep{kupiec1995techniques} to backtest the model.  We use $W=100$ train windows and train the model on each, starting from randomly initialized weights. Denote by $X_h^{s,w}$ the ground truth target value for series $s\in\{1,\ldots, S\}$, prediction horizon $h\in\{1,\ldots, H\}$, and training window $w\in\{1,\ldots, W\}$. Let $\alpha$ be the confidence level and denote the corresponding quantile estimate of the model for the $h$-step ahead prediction in window $w$ and series $s$, by $\hat{q}_{h}^{s,w}(\alpha)$. Then, the number of model failures to correctly predict the quantile can be measured by a statistics $\hat{v}_{h,\alpha}^s$ defined as:
\[
    \hat{v}_{h,\alpha}^s=\sum_{w=1}^{W}\mathbf{1}(\hat{q}_{h}^{s,w}(\alpha) < X_{h}^{s,w}).
\]
If the model is correct (that is, under the null hypothesis), $\hat{v}_{h,\alpha}^s$ has the Binomial distribution with parameters $(1-\alpha, W)$, see \citep[Section 9.3.1]{mcneil2015quantitative}. The likelihood ratio test is given by 
\[
    T_{h,\alpha}^s = 2\log Bin(\hat{v}_{h,\alpha}^a, \hat{f}_{h,\alpha}^s, W) - 2\log Bin(\hat{v}_{h,\alpha}^s,1-\alpha, W),
\]
where $\hat{f}_{h,\alpha}\triangleq\hat{v}_{h,\alpha}^s/W$ and $Bin(k,p,n) = {n \choose k}p^k (1-q)^{n-k}$. Under the null hypothesis, $T_{h, \alpha}^s$ is asymptotically distributed as $\chi$-squared distribution with one degree of freedom, $\chi^2(1)$.

Correspondingly, the $p$-value is computed as 
\[
    p\text{-value}_{h,\alpha}^s = \overline{F}_{\chi^2(1)}(T_{h, \alpha}^s),
\]
where $F_{\chi^2(1)} = 1 - \overline{F}_{\chi^2(1)}$ is the cumulative distribution function of $\chi^2(1)$. We report a proportion of all time series for which the p-value is at least $\gamma$, i.e., 
\[
    p\text{-value}_{\alpha} = \frac{1}{SH}\sum_{h=1}^H\sum_{s=1}^S\mathbf{1}(p\text{-value}_{h, \alpha}^s\ge \gamma),
\]
where $S$ denotes the number of time series in the dataset. 

\section{Computational infrastructure, cost, and source code}\label{app:infra} 

We conducted our experiments using a cluster with Nvidia A100 graphics cards. Each training used a single GPU card, up to 128GB of RAM, and 16 CPU cores. Training our model took between 30 minutes and 2 hours, depending on the number of time series in the dataset. The simulation phase (when we sampled 1024 predictions) took up to $3.5$ hours. We estimate the overall cost of reproducing the results to be $2000$ GPUh.

The project's overall cost, including development, prototyping, and evaluation of our method and benchmarking the baseline methods, took approximately $80$K GPUh.

\section{The Quick, Draw! Dataset}
\label{app:hand-drawn-images}

To check how our method performs on tasks, which do not involve `traditional' time series, we trained \method{} on apple sketches of The Quick, Draw! Dataset\footnotemark{}. This dataset is a collection of $50$ million drawings divided into $345$ categories, gathered from players of the game Quick, Draw!. In this experiment, we treat sketches as two-dimensional $(x,y)$ time series, an extension to our univariate input and output discretization schema from Section~\ref{sec:method}. 
\footnotetext{\url{https://github.com/googlecreativelab/quickdraw-dataset}}

When modelling a $d$-variate time series, we consider $\mathbb X_{t-T+1:t}=(X_{t-T+1}, \ldots, X_{t})$, where now $X_t\in \mathbb R^d$. A time series $\mathbb{X}$ is translated into a sequence of tokens, by first flattening it into a one-dimensional sequence of intertwined series, which has a length $d\cdot T$. Each value in such sequence is then broken down into digits, as described in Section~\ref{sec:method}. This results in a series of tokens $\tau(\mathbb{X})=(\tau_t(\mathbb{X}))_{t=1}^{p\cdot d\cdot T}$ of length $p\cdot d\cdot T$. After processing it with \method{}, tokens are generated one by one and gathered into groups of size $p \cdot d$ which are later reinterpreted as distinct time steps of a multivariate series. 

\begin{figure}[htbp]
    \centering
    \includegraphics[width=0.7\textwidth]{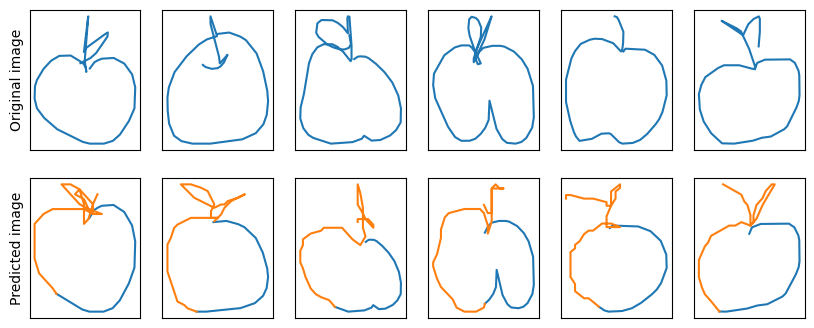}
    \caption{Reconstruction of apple sketches by \method{}. The ground truth images are shown in the top row. In the bottom row, the partial ground truth trajectory used to prompt \method{} is shown in blue, and the reconstructed trajectory is highlighted in orange.}
    \label{fig:apple_sketches}
\end{figure}

\begin{figure}[htbp]
    \centering
    \includegraphics[width=0.7\textwidth]{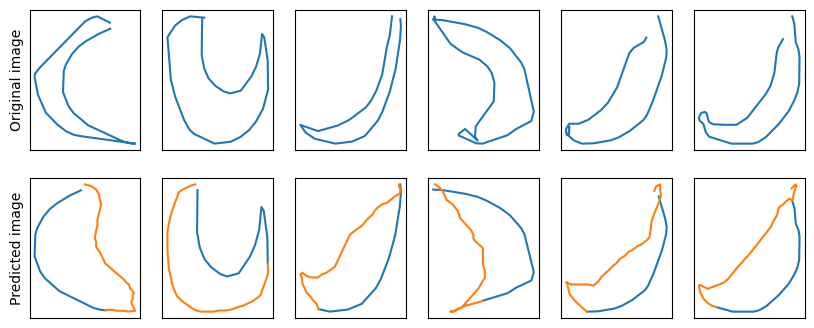}
    \caption{Reconstruction of banana sketches by \method{}. The ground truth images are shown in the top row. In the bottom row, the partial ground truth trajectory used to prompt \method{} is shown in blue, and the reconstructed trajectory is highlighted in orange.}
    \label{fig:app:banana}
\end{figure}

\begin{figure}[htbp]
    \centering
    \includegraphics[width=0.9\textwidth]{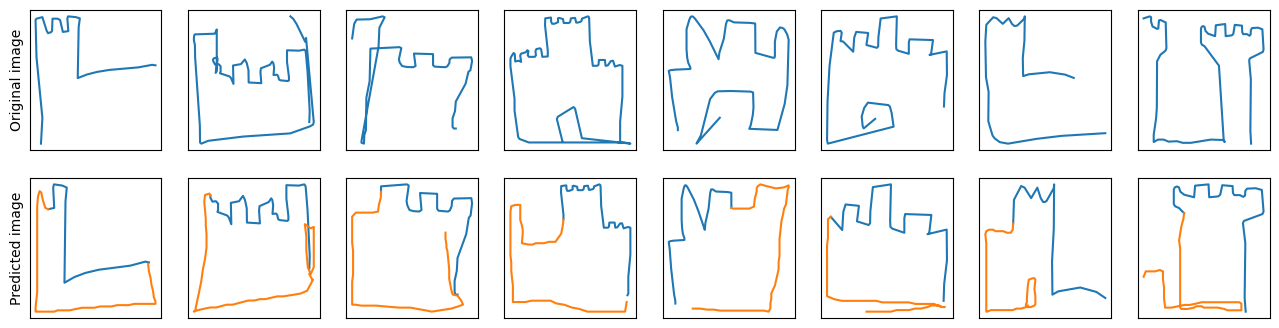}
    \caption{Reconstruction of castle sketches by \method{}. The ground truth images are shown in the top row. In the bottom row, the partial ground truth trajectory used to prompt \method{} is shown in blue, and the reconstructed trajectory is highlighted in orange.}
    \label{fig:app:castle}
\end{figure}

\begin{figure}[htbp]
    \centering
    \includegraphics[width=0.95\textwidth]{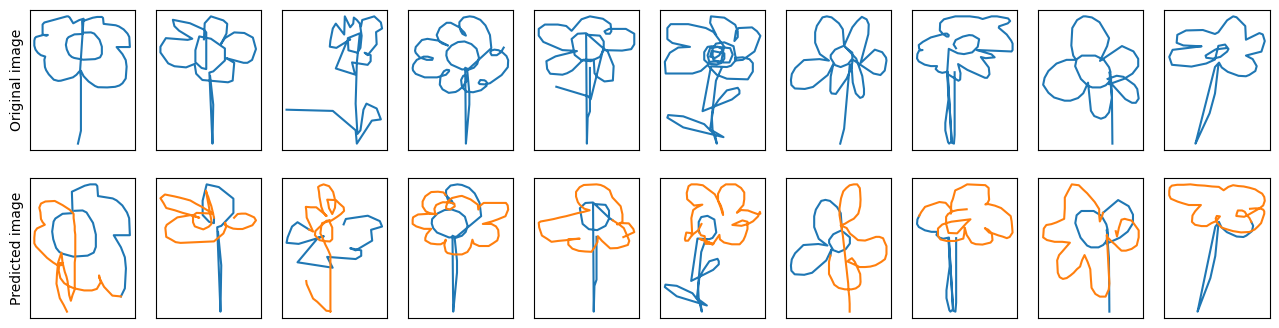}
    \caption{Reconstruction of flower sketches by \method{}. The ground truth images are shown in the top row. In the bottom row, the partial ground truth trajectory used to prompt \method{} is shown in blue, and the reconstructed trajectory is highlighted in orange.}
    \label{fig:app:flower}
\end{figure}

In evaluation, we remove and reconstruct a part of the image with the model. Selected examples are shown in Figure \ref{fig:apple_sketches}. We observe that, in general, the reconstructions are plausible and faithfully depict high-level features (e.g., the shape is 'closed').  On the negative side, the model struggles with low-level details. Samples of \method{} reconstructing images from the banana, castle, or flower classes are given in Figure \ref{fig:app:banana}, Figure \ref{fig:app:castle}, and Figure \ref{fig:app:flower}, respectively.

\section{Ablations}\label{app:ablations}
In this section, we present several ablation studies. We analyze different design choices for scaling the raw time series input, normalization to $[0, 1]$, embeddings, and significance weighting. We conducted the experiments on \electricity{} dataset.

\subsection{Scaling the raw time series input}

Before data is fed to the transformer, it is scaled. \method{} performs that by dividing the input sequence of length $T$ by its mean (following \cite{li2019enhancing}). This approach, however, leaks data in the sense that each token is scaled by the statistic dependent on the whole context. Theoretically, the model could exploit this information during training at the expense of its performance at prediction time when it suffers a distribution shift by only having access to the mean of $T-H$ values. To check whether this matters in practice, we implemented a causal normalization procedure, where each value in the sequence is normalized only by the mean of the preceding input prefix. Using a well-known recursive property of rolling mean avoids information leaks and, consequently, distribution shift at inference time. We present the results of the comparison in Table~\ref{tab:ablations_norm}. Somewhat surprisingly, the only significant difference between the approaches is on QL-99, where the non-causal approach achieves better results. This might suggest that the data leak problem is not too severe or that the other data preprocessing steps mitigate the problem. The non-causal approach used in \method{} is also conceptually simpler and easier to code.

\begin{table*}[htpb]
    \centering
    \caption{Comparison of different normalization methods on \electricity{} dataset. The presented values are Inter Quantile Means computed over 10 training windows and  321 training time series. 95\% bootstrap confidence interval is provided in brackets.}
    \label{tab:ablations_norm}
    \begin{tabular}{lllll}
        \toprule
        model &                        RMSE &                            MAD &                              QL-99 &                            QL-95 \\
        \midrule
        \method{} per-ts norm. & 0.062\tiny{(0.058, 0.067)} & 0.047\tiny{(0.044, 0.050)} & 0.004\tiny{(0.004, 0.004)} & 0.014\tiny{(0.013, 0.015)}\\
        \method{} causal norm. & 0.063\tiny{(0.058, 0.068)} & 0.048\tiny{(0.044, 0.051)} & 0.005\tiny{(0.004, 0.005)} & 0.015\tiny{(0.014, 0.016)}\\
        \bottomrule
    \end{tabular}
\end{table*}

\subsection{Normalization to $[0, 1]$}

After normalization, we scale the input data to $[0, 1]$. \method{} follows \citet{li2019enhancing} in reducing the range of data by dividing by a constant factor ($10$ in our case) and clipping the outcome to $[0, 1]$. Alternatively, we tested a sigmoid function for softer scaling. Intuitively, this approach could preserve outliers better as it provides additional space for samples far from the average value. The results of the comparison can be found in Table \ref{tab:serialization_ablation}. The metric values are higher for the approach used by \method{}, though the approaches are comparable in statistical terms. 

\begin{table*}[htbp]
    \centering
    \caption{Comparison of different preprocessing methods on \electricity{} dataset. The presented values are Inter Quantile Means computed over 10 training windows and  321 training time series. 95\% bootstrap confidence interval is provided in brackets.}
    \label{tab:serialization_ablation}
    \begin{tabular}{lllll}
        \toprule
        model &                        RMSE &                            MAD &                              QL-99 &                            QL-95 \\
        \midrule
        \method{} clip & 0.062\tiny{(0.058, 0.067)} & 0.047\tiny{(0.044, 0.050)} & 0.004\tiny{(0.004, 0.004)} & 0.014\tiny{(0.013, 0.015)}\\
        \method{} squash & 0.064\tiny{(0.059, 0.069)} & 0.048\tiny{(0.045, 0.052)} & 0.004\tiny{(0.004, 0.004)} & 0.014\tiny{(0.013, 0.015)}\\
        \bottomrule
    \end{tabular}
\end{table*}

\subsection{Significance weighting $\beta$}
To train \method{}, we use a weighted loss, see \eqref{eq:loss}. The parameter $\beta$ controls the contribution of each digit to the loss according to its significance. We compare the performance of different values of $\beta$, see Table \ref{tab:beta_ablation}. In most cases, the metrics for $\beta=0.3$ (used in \method{}) are the most favorable, though from a statistical perspective, the outputs are rather stable under different choices of $\beta$.

\begin{table*}[htbp]
    \centering
    \caption{Comparison of different $\beta$ values on \electricity{} dataset. The presented values are Inter Quantile Means computed over 10 training windows and  321 training time series. 95\% bootstrap confidence interval is provided in brackets.}
    \label{tab:beta_ablation}
    \begin{tabular}{lllll}
        \toprule
        model &                        RMSE &                            MAD &                              QL-99 &                            QL-95 \\
        \midrule
        \method{} $\beta$=0.3 & 0.062\tiny{(0.058, 0.067)} & 0.047\tiny{(0.044, 0.050)} & 0.004\tiny{(0.004, 0.004)} & 0.014\tiny{(0.013, 0.015)}\\
        \method{} $\beta$=0.6 & 0.065\tiny{(0.061, 0.071)} & 0.050\tiny{(0.046, 0.054)} & 0.004\tiny{(0.004, 0.005)} & 0.015\tiny{(0.014, 0.016)}\\
        \method{} $\beta$=0.9 & 0.064\tiny{(0.059, 0.069)} & 0.048\tiny{(0.045, 0.052)} & 0.004\tiny{(0.004, 0.004)} & 0.014\tiny{(0.013, 0.015)}\\
        \method{} $\beta$=0.99 & 0.065\tiny{(0.061, 0.069)} & 0.049\tiny{(0.047, 0.052)} & 0.004\tiny{(0.003, 0.004)} & 0.014\tiny{(0.013, 0.015)}\\
        \bottomrule
    \end{tabular}
\end{table*}

\subsection{Discretization and embeddings}

Normalized real values are discretized before being passed to \method{}, as described in Section \ref{app:normalization}. When we fix the normalization factor, the function that assigns the most significant digit to raw values splits the space of raw inputs $\mathbb{R}$ into $B$ regions. By the linearity of transformations, those regions, apart from the two outermost ones, form intervals of the same length. We could consider other discretization schemes, which assign the most significant digits differently. In particular, we consider a function which splits the raw input space into $B$ uneven intervals, such that those intervals contain a similar number of observed values. Intuitively, instead of evenly splitting the input space, we split evenly along the quantiles of empirical distribution of time series values. We present the results of this ablation in Table \ref{tab:emb_ablation}. On the two datasets on which we run this ablation, we find no advantage to using this method.

Our model uses the same tokens to represent digits of different significance. Potentially, using the exact same embeddings coupled with the distributional differences between tokens appearing on different positions could impact models performance. We check, if allowing \method{} to learn different embeddings for digits appearing on different positions could improve its performance, but find that this is not the case, as evident from Table \ref{tab:emb_ablation}. We hypothesize that out method is able to capture temporal dependencies well enough to distinguish the position in which a digit appears and builds common internal representation of numbers.

\begin{table*}
    \centering
    \caption{
    Comparison of different approaches to discretization and digit embedding. ``Quantile discretization'' refers to a discretization schema in which the most significant digits appear with the similar frequency.  ``Separate embeddings'' refers to the experiment in which the same digits appearing at different positions in a representation of a number are assigned different vectors in the embedding space.
    }
    \label{tab:emb_ablation}
    \begin{tabular}{lllll}
        \toprule
        metric & model & \electricity{} & \ettm{}\\
        \midrule
        MAD & \method{} -- quantile discretization & 0.060\tiny{(0.059, 0.060)} & 0.095\tiny{(0.088, 0.102)}\\
        & \method{} -- separate embeddings & 0.062\tiny{(0.061, 0.062)} & 0.092\tiny{(0.087, 0.099)}\\
        & \method{} & 0.057\tiny{(0.057, 0.058)} & 0.092\tiny{(0.086, 0.098)} \\
        \midrule
        RMSE & \method{} -- quantile discretization & 0.078\tiny{(0.077, 0.078)} & 0.116\tiny{(0.107, 0.125)}\\
        & \method{} -- separate embeddings & 0.080\tiny{(0.079, 0.080)} & 0.114\tiny{(0.107, 0.123)}\\
        & \method{} & 0.075\tiny{(0.074, 0.075)} & 0.113\tiny{(0.105, 0.121)} \\
        \bottomrule
    \end{tabular}
\end{table*}

\end{document}

%% file: math_commands.tex
\usepackage{amsmath,amsfonts,bm}

\def\eqref#1{equation~\ref{#1}}

\def\1{\bm{1}}

\DeclareMathAlphabet{\mathsfit}{\encodingdefault}{\sfdefault}{m}{sl}
\SetMathAlphabet{\mathsfit}{bold}{\encodingdefault}{\sfdefault}{bx}{n}

